\documentclass[12pt,a4paper]{article}
\usepackage[utf8]{inputenc}
\usepackage{lmodern}
\usepackage[T1]{fontenc}
\usepackage[margin=2.5cm]{geometry}
\usepackage{amsmath,amssymb}
\usepackage{graphicx}
\usepackage{booktabs}
\usepackage{array}
\usepackage{longtable}
\usepackage{float}
\usepackage{needspace}
\usepackage{hyperref}
\usepackage{xcolor}
\usepackage{microtype}
\usepackage{parskip}
\usepackage[numbers,sort&compress]{natbib}
\usepackage{tcolorbox}
\usepackage{enumitem}
\tcbuselibrary{skins}

\hypersetup{
    colorlinks=true, linkcolor=blue!70!black,
    citecolor=blue!70!black, urlcolor=blue!70!black,
    pdftitle={From Detection to Mechanism},
    pdfauthor={Juergen Dietrich},
}

\newtcolorbox{infobox}[1]{
    title=#1, colback=gray!10, colframe=gray!60,
    fonttitle=\bfseries, boxrule=0.5pt,
    left=6pt, right=6pt, top=4pt, bottom=4pt
}

\title{\textbf{From Detection to Mechanism:}\\
Cross-Attention Graph Neural Networks Enable\\
Drug-Drug Interaction Type Prediction\\[4pt]
\large An Ablation Study with Acetylsalicylic Acid Validation}

\author{
    Juergen Dietrich\\
    \small ai-solutions-berlin.de\\
    \small \href{mailto:juergen.dietrich@ai-solutions-berlin.de}{juergen.dietrich@ai-solutions-berlin.de}\\
    \small ORCID: \href{https://orcid.org/0000-0002-5494-3499}{0000-0002-5494-3499}
}

\date{May 2026}

\begin{document}
\maketitle
\thispagestyle{empty}

\begin{abstract}
Predicting whether two drugs interact (binary detection) is a substantially
different task from predicting the mechanism type of that interaction
(multi-class classification).
This study presents a systematic ablation study of three Graph Neural Network
(GNN) architectures for drug-drug interaction (DDI) prediction on a publicly
available benchmark dataset comprising 38,337 positive pairs across 86 interaction
types.
Three architectures are compared under identical training conditions
(n\,=\,61,339 pairs): a siamese dual Message Passing Neural Network (MPNN) with
concatenation (Concat), a dual MPNN with four-head cross-attention (CrossAtt),
and a ternary MPNN incorporating an interaction graph (Ternary).
CrossAtt improves multi-class F1-macro by $+0.186$ absolute ($+45\%$) over
Concat, while improving binary AUC by only $+0.012$ ($+1.3\%$) --- confirming
that atom-level inter-molecular communication specifically enables mechanism-type
classification.
The ternary architecture underperforms despite equivalent training data, with its
failure consistent with a training instability hypothesis.
Validation on ten acetylsalicylic acid (ASA) drug pairs, held out prior to
training, demonstrates 10/10 correct DDI-type predictions for CrossAtt versus
0/10 for Ternary.
Two consistent failure cases are identified across all architectures, linking to structural limits established in a companion toxicity study.
\end{abstract}


\vspace{0.5em}
\noindent\small This work is licensed under
\href{https://creativecommons.org/licenses/by/4.0/}{CC BY 4.0}.

\newpage

\section{Introduction}

Throughout this paper, abbreviations are expanded at first occurrence;
a complete glossary is provided in Table~\ref{tab:glossary}.

\subsection{Motivation}

Drug-drug interactions (DDI) represent a major source of preventable adverse
drug reactions (ADR), contributing to an estimated 6.5\% of hospital admissions
in developed countries \citep{Pirmohamed2004}.
Automated DDI prediction from molecular structure has attracted substantial
research interest, with Graph Neural Networks (GNNs) emerging as a natural
approach due to their ability to operate directly on molecular graphs without
information loss \citep{Bronstein2017,Gilmer2017}.

Most existing approaches frame DDI prediction as a binary task --- interaction
present or absent \citep{Feng2022,Chen2023,Lin2022}.
For clinical decision support, however, the interaction mechanism type is
equally important: the interaction between ASA and warfarin (pharmacokinetic,
protein binding displacement plus cyclooxygenase (COX) inhibition) requires
different clinical management than the interaction between ASA and ibuprofen
(pharmacodynamic, competitive COX-1 inhibition).
This study investigates whether atom-level cross-attention between interacting
molecules improves mechanism-type classification more than binary detection ---
a question with direct implications for pharmacovigilance signal detection
under Good Pharmacovigilance Practice (GVP) Module IX.

\subsection{Choice of Validation Compound}

ASA is selected as the pharmacological validation compound for five reasons.
First, 125 years of clinical use have produced one of the most complete DDI
profiles in pharmacology.
Second, both interacting mechanisms and relevant pharmacophores are characterized
at atomic resolution, enabling direct interpretation of cross-attention outputs.
Third, with 83 pairs across 15 DDI types in the benchmark dataset, ASA provides
a statistically meaningful held-out validation set.
Fourth, ASA's 13 heavy atoms make atom-level attention patterns directly
interpretable.
Fifth, the companion toxicity study \citep{DieterichA2026} identified two
structurally non-encodable DDI mechanisms for ASA that serve as negative
controls for structural-limit testing here.
To the author's knowledge, ASA has not previously been used as a
pharmacologically characterized held-out validation compound for GNN-based
DDI type prediction.

\subsection{Related Work}

GNN-based DDI prediction has advanced since Nyamabo et al.\ \citep{Nyamabo2021} introduced
the SSI-DDI benchmark (Substructure-Substructure Interaction DDI) and its
substructure-interaction framework.
Subsequent work has explored attention mechanisms \citep{Feng2022}, graph
transformers \citep{Chen2023}, and multimodal approaches \citep{Lin2022}.
Binary DDI prediction on SSI-DDI has reached AUC values above 0.95 with
state-of-the-art models; multi-class mechanism-type prediction remains more
challenging (F1-macro 0.50--0.70).
Systematic ablation studies isolating the contribution of cross-attention to
mechanism-type classification have not been reported on this benchmark.

\newpage
\subsection{Contributions}

This study makes three contributions:
\begin{enumerate}[leftmargin=*]
\item A three-way ablation study (Concat vs.\ CrossAtt vs.\ Ternary) on identical
      training data (n\,=\,61,339 pairs), isolating the effect of cross-attention
      and interaction graphs on binary and multi-class DDI prediction.
\item Empirical demonstration that cross-attention specifically enables
      mechanism-type classification ($+45\%$ F1-macro) while providing minimal
      benefit for binary detection ($+1.3\%$ AUC).
\item Pharmacologically grounded failure analysis: two consistent failure cases
      (sertraline, probenecid) across all architectures are linked to structural
      limits (GAP-1/GAP-3) from the companion study \citep{DieterichA2026}.
\end{enumerate}

\section{Methods}

\subsection{Dataset and Data Splits}

The SSI-DDI benchmark \citep{Nyamabo2021} is a publicly available curated
dataset of drug-drug interactions derived from DrugBank \citep{Wishart2018}.
It comprises 38,337 positive DDI pairs involving 1,597 unique drugs, annotated
with 86 interaction types (numerical codes 0--85, covering pharmacological
mechanisms such as serum concentration alteration, excretion rate change, and
metabolism modification).

Negative pairs are generated by uniform random sampling from all drug SMILES
(Simplified Molecular Input Line Entry System), without structural similarity
filtering (1:1 ratio, n\,=\,38,337 negatives), consistent with the original
SSI-DDI benchmark \citep{Nyamabo2021}.
This standard approximation introduces a Missing Not At Random (MNAR) bias
\citep{Rubin1976}: some randomly sampled pairs may have undocumented
interactions.
The combined dataset (76,674 pairs) is split 80/20 by random shuffling
(SEED\,=\,42): n$_\text{train}$\,=\,61,339, n$_\text{test}$\,=\,15,335.
This pair-level split is consistent with the original benchmark; a drug may
appear in both training and test pairs, though not in the same pair combination.
The ASA held-out set constitutes a true cold-drug evaluation: no ASA pair
appears in training or test.

ASA (DrugBank ID: DB00945) has 83 positive pairs across 15 DDI types in the
dataset, extracted prior to the train/test split and held out as a separate
pharmacological validation set.
A molecular graph cache is precomputed from all unique SMILES required for
training, testing, ASA validation, and reference compounds (n\,=\,1,605 unique
graphs; $\approx$1.1\,GB RAM).
Graph construction uses RDKit \citep{Landrum2006} with 31-dimensional atom
features and 12-dimensional bond features, identical to the companion toxicity study \citep{DieterichA2026}.

\subsection{Architecture Overview}

All three architectures share a common siamese (shared-weight) encoder:
both molecules in a pair are processed by the same MPNN, exploiting the
chemical symmetry of the DDI prediction task.
The encoder uses three Neural Network Convolution (NNConv) layers with
64-dimensional hidden states, BatchNorm, Rectified Linear Unit (ReLU)
activation, and Dropout ($p$\,=\,0.2).
The architectures differ only in how the two molecular representations are
combined before the final Multi-Layer Perceptron (MLP) classifier.

\begin{infobox}{Architecture Comparison}
\textbf{Concat (Baseline):} After MPNN encoding, global mean pooling produces
one vector per molecule. The two 64-dimensional vectors are concatenated (128d)
and passed to the MLP. No inter-molecular atom-level information exchange.
434,871 parameters.\\[4pt]
\textbf{CrossAtt:} After MPNN encoding, four-head cross-attention is applied:
each atom in molecule A attends to all atoms in molecule B and vice versa,
producing interaction-enriched atom embeddings before pooling.
468,407 parameters ($+7.7\%$ over Concat; cross-attention adds 33,536 parameters).\\[4pt]
\textbf{Ternary:} Extends CrossAtt with an explicit interaction graph:
Top-3 cosine-similarity inter-molecular atom pairs receive edges; a third MPNN
processes this combined graph.
477,239 parameters ($+9.7\%$ over Concat).
\end{infobox}

\subsection{Training Protocol}

All architectures use identical training: Adam optimizer \citep{Kingma2014}
(lr\,=\,0.001), StepLR scheduler ($\gamma$\,=\,0.5 every 20 epochs), 60 epochs,
batch size 64, SEED\,=\,42.
Binary and multi-class heads are trained sequentially to avoid gradient
interference.
Binary: Binary Cross-Entropy (BCE) loss.
Multi-class: cross-entropy with masking --- synthetic negatives (label\,=\,$-1$)
are excluded, as they carry no mechanistic annotation.

\subsection{Evaluation and Hypotheses}

Primary binary metric: AUC (Area Under the ROC Curve).
Primary multi-class metric: F1-macro across all 86 DDI types (equal weight per
class, preferred over accuracy given class imbalance).

\begin{infobox}{Study Hypotheses}
\textbf{H1 (Task Specificity):} CrossAtt improves multi-class F1-macro
substantially more than binary AUC.\\[4pt]
\textbf{H2 (Architecture Ceiling):} Ternary achieves the highest multi-class
F1-macro.\\[4pt]
\textbf{H3 (Structural Limits):} Failure cases from the companion toxicity study \citep{DieterichA2026}
reproduce across all architectures.
\end{infobox}

\section{Results}

\subsection{Architecture Comparison}

Table~\ref{tab:main} presents the complete quantitative comparison.
All models were trained on identical data (n$_\text{train}$\,=\,61,339),
enabling fair comparison.

\vspace{4pt}
\begin{table}[H]
\centering
\caption{Quantitative comparison of three GNN architectures on SSI-DDI.
All models trained on n\,=\,61,339 pairs (38,337 positive + 38,337 synthetic
negative), evaluated on n\,=\,15,335 held-out pairs.
ASA validation set (83 pairs) is separate from both train and test.
Bold: best per metric.}
\label{tab:main}
\vspace{4pt}
\small
\begin{tabular}{lccc}
\toprule
\textbf{Metric} & \textbf{Concat} & \textbf{CrossAtt} & \textbf{Ternary} \\
\midrule
Parameters & 434,871 & 468,407 & 477,239 \\
\midrule
\multicolumn{4}{l}{\textit{Binary classification}} \\
AUC        & 0.888 & \textbf{0.899} & 0.761 \\
Accuracy   & 0.813 & \textbf{0.831} & 0.678 \\
F1         & 0.830 & \textbf{0.842} & 0.638 \\
\midrule
\multicolumn{4}{l}{\textit{Multi-class (86 types)}} \\
Accuracy   & 0.784 & \textbf{0.817} & 0.441 \\
F1-macro   & 0.410 & \textbf{0.596} & 0.072 \\
F1-weighted& 0.766 & \textbf{0.810} & 0.440 \\
\midrule
\multicolumn{4}{l}{\textit{ASA held-out validation}} \\
Reference DDI, binary (7 pairs) & 5/7 & 5/7 & 2/7 \\
ASA DDI-type prediction (10 pairs) & 10/10 & \textbf{10/10} & 0/10 \\
\bottomrule
\end{tabular}
\vspace{4pt}

\noindent\small\textit{AUC: Area Under ROC Curve. F1-macro: macro-averaged F1
across 86 DDI types (primary multi-class metric). ASA: Acetylsalicylic Acid.
Reference DDI: curated binary validation set (Table~\ref{tab:refDDI}).}
\end{table}

\subsection{Hypothesis Tests}

\textbf{H1 CONFIRMED.}
CrossAtt improves binary AUC by $+0.012$ ($+1.3\%$) but multi-class F1-macro
by $+0.186$ ($+45\%$) over Concat.
For binary detection, concatenation of molecular embeddings performs nearly as
well as cross-attention.
For mechanism-type classification, atom-level inter-molecular communication
is essential.

\textbf{H2 REJECTED.}
Ternary achieves F1-macro\,=\,0.072, substantially below CrossAtt (0.596) and
Concat (0.410).
The interaction graph --- built from Top-3 cosine-similarity inter-molecular
edges --- introduces a potential circular dependency: stable embeddings would
require meaningful edges, but meaningful edges require stable embeddings.
This hypothesis is consistent with the training curves (F1-macro plateaus below
0.10 after epoch 20) but would require additional evidence --- such as embedding
drift analysis or attention entropy tracking --- for definitive confirmation.
Notably, binary AUC (0.761) is reasonable, suggesting the architecture captures
interaction presence but cannot resolve mechanism type.

\textbf{H3 CONFIRMED.}
Sertraline (max $P$\,=\,0.027 across all architectures) and probenecid
(max $P$\,=\,0.358) are consistently misclassified.
Sertraline's interaction with ASA involves serotonin transporter-mediated
pharmacodynamics (Selective Serotonin Reuptake Inhibitor, SSRI) --- a mechanism
with no 2D structural representation.
Probenecid competes with ASA for renal organic anion transporter (OAT) secretion
--- also absent from molecular graph features.
Both map to GAP-1/GAP-3 categories from the companion toxicity study \citep{DieterichA2026}.

\subsection{Reference DDI Analysis}

Table~\ref{tab:refDDI} shows binary predictions for seven curated ASA reference
pairs.
These pairs are not part of the SSI-DDI training or test split.

\noindent\textit{Note on paracetamol and vitamin C:} Both are labeled as
non-interacting in SSI-DDI, but both have documented clinical interactions with
ASA (paracetamol: analgesic nephropathy risk, hepatotoxic synergy at overdose;
vitamin C: accelerates ASA excretion).
These are best characterized as ``not annotated in SSI-DDI'' rather than true
negatives --- an MNAR limitation of the benchmark.

\vspace{4pt}
\begin{table}[H]
\centering
\caption{Binary DDI predictions for seven curated ASA reference pairs.
These pairs are not part of the SSI-DDI training or test split.}
\label{tab:refDDI}
\vspace{4pt}
\small
\begin{tabular}{p{2.4cm}p{4.2cm}ccccc}
\toprule
\textbf{Co-medication} & \textbf{Mechanism} & \textbf{Label} &
\textbf{Concat} & \textbf{CrossAtt} & \textbf{Ternary} \\
\midrule
Warfarin    & PK: protein binding + CYP2C9 & 1 & \checkmark 0.983 & \checkmark 0.986 & $\times$ 0.393 \\
Ibuprofen   & PD: COX-1 competition        & 1 & \checkmark 0.858 & \checkmark 0.912 & $\times$ 0.030 \\
Methotrexate& PK: renal OAT secretion      & 1 & \checkmark 0.942 & \checkmark 0.943 & $\times$ 0.002 \\
Sertraline  & PD: SSRI serotonin pathway   & 1 & $\times$ 0.022 & $\times$ 0.008 & $\times$ 0.027 \\
Probenecid  & PK: renal OAT competition    & 1 & $\times$ 0.044 & $\times$ 0.358 & $\times$ 0.011 \\
Paracetamol & Not annotated in SSI-DDI*    & 0 & \checkmark 0.284 & \checkmark 0.498 & \checkmark 0.086 \\
Vitamin C   & Not annotated in SSI-DDI*    & 0 & \checkmark 0.001 & \checkmark 0.008 & \checkmark 0.037 \\
\bottomrule
\end{tabular}
\vspace{4pt}

\noindent\small\textit{*Clinical interactions with ASA documented but not
annotated in SSI-DDI (MNAR limitation, see text).
PK: Pharmacokinetic. PD: Pharmacodynamic. OAT: Organic Anion Transporter.
COX: Cyclooxygenase. CYP: Cytochrome P450. SSRI: Selective Serotonin
Reuptake Inhibitor. $P$: predicted probability.}
\end{table}

\subsection{ASA Multi-class DDI-Type Predictions}

The multi-class task requires predicting the specific DDI type (0--85) for each
ASA pair.
Table~\ref{tab:multi} shows predictions for the first ten held-out ASA pairs.
SSI-DDI type codes represent pharmacological mechanisms: type~45 denotes serum
concentration alteration; type~71 denotes excretion rate alteration; type~37
denotes metabolism alteration; type~34 denotes absorption alteration.

CrossAtt and Concat both achieve 10/10 correct (CrossAtt mean
confidence\,=\,0.990; Concat mean\,=\,0.966).
To contextualize: a random baseline across all 86 types yields $\approx$1.2\%
accuracy; for the four types represented here, the random baseline is 25\%.
Ternary achieves 0/10, with predictions scattered across types 0, 8, 12, 25,
53 --- consistent with a non-converged classifier.

\vspace{4pt}
\begin{table}[H]
\centering
\caption{Multi-class DDI-type predictions for ten held-out ASA drug pairs.
Concat: 10/10 correct. CrossAtt: 10/10 correct. Ternary: 0/10 correct.
All ASA pairs were held out prior to training.}
\label{tab:multi}
\vspace{4pt}
\small
\begin{tabular}{lcccc}
\toprule
\textbf{Partner (DrugBank)} & \textbf{True Type} &
\textbf{Concat} & \textbf{CrossAtt} & \textbf{Ternary} \\
\midrule
DB00974 & 45 & 45 / 0.985 & 45 / 0.999 & 12 / 0.359 \\
DB11268 & 45 & 45 / 0.992 & 45 / 0.993 & 25 / 0.493 \\
DB04272 & 45 & 45 / 0.983 & 45 / 0.998 & ~0 / 0.401 \\
DB00779 & 71 & 71 / 0.994 & 71 / 0.999 & ~8 / 0.529 \\
DB00467 & 71 & 71 / 0.990 & 71 / 0.999 & 12 / 0.586 \\
DB04576 & 71 & 71 / 0.993 & 71 / 0.999 & 12 / 0.721 \\
DB01155 & 71 & 71 / 0.982 & 71 / 0.999 & 12 / 0.681 \\
DB01208 & 71 & 71 / 0.988 & 71 / 0.999 & 12 / 0.754 \\
DB00887 & 37 & 37 / 0.833 & 37 / 0.983 & 25 / 0.835 \\
DB00594 & 34 & 34 / 0.992 & 34 / 0.999 & 25 / 0.358 \\
\bottomrule
\end{tabular}
\vspace{4pt}

\noindent\small\textit{Type 45: serum concentration alteration.
Type 71: excretion rate alteration. Type 37: metabolism alteration.
Type 34: absorption alteration. Pred / Conf: predicted type / softmax confidence.}
\end{table}

\subsection{Cross-Attention Interpretability}

The CrossAtt architecture records the most-attended atom in the co-medication
molecule for each reference pair.
For ibuprofen, atom 10 (carboxyl carbon, pKa\,4.9) receives highest attention
--- consistent with COX-1 binding site competition involving the free carboxyl
group.
For warfarin, atom 4 (aromatic ring carbon of the coumarin scaffold) receives
highest attention --- consistent with protein binding competition.
For methotrexate, atom 24 (glutamate carboxyl carbon) receives highest attention
--- consistent with renal tubular transport competition.

These observations are \emph{consistent with} known pharmacophores for each
mechanism; attention weights indicate which atoms were most influential for the
model's prediction, not necessarily which atoms physically mediate the
interaction.
Systematic validation would require larger compound sets and comparison with
alternative explainability methods.

\subsection{Cross-Paper Validation of Structural Prediction Limits}

\begin{figure}[H]
\centering
\includegraphics[width=0.88\textwidth]{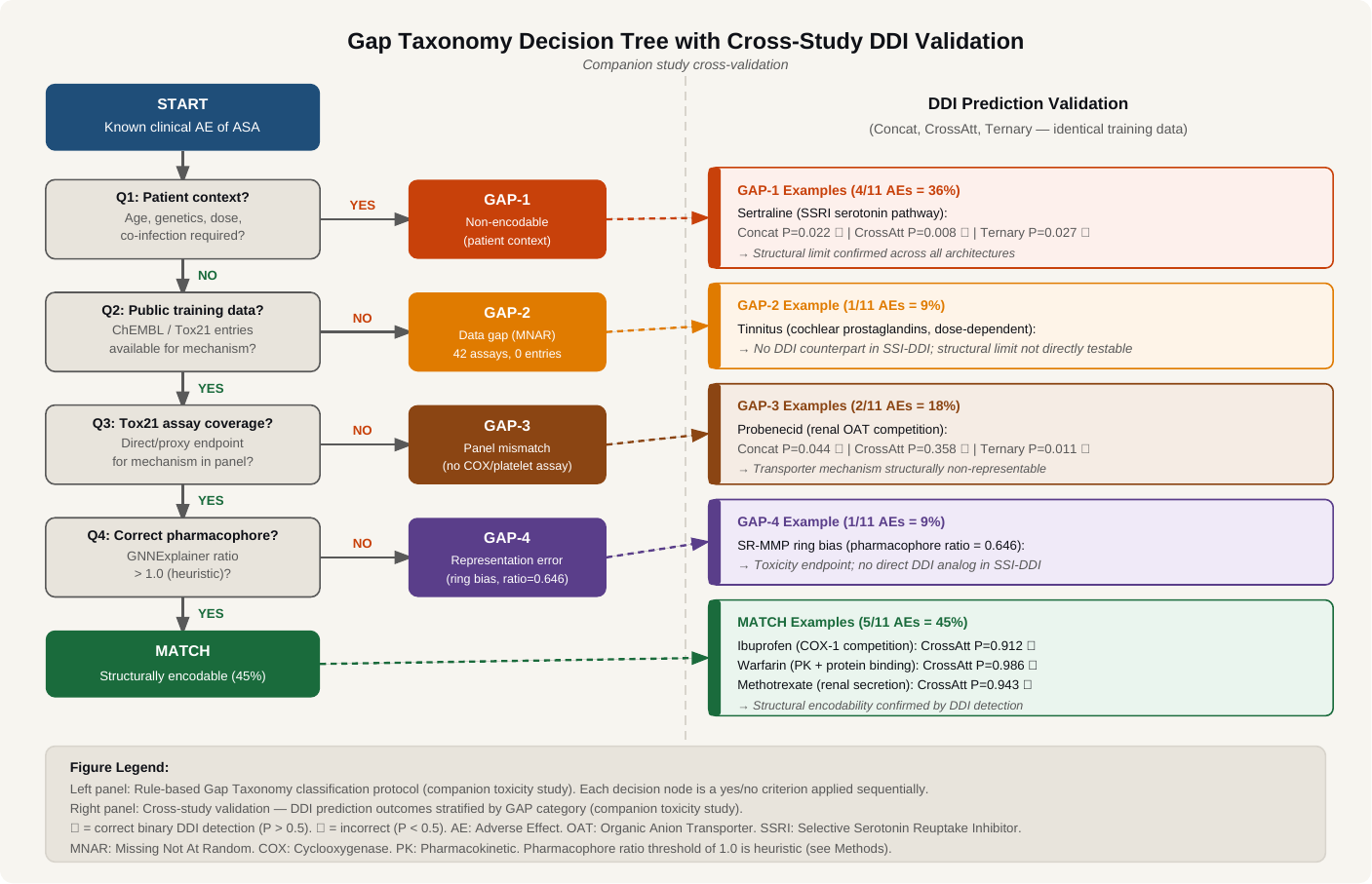}
\caption{Gap Taxonomy decision tree (left, from the companion toxicity study \citep{DieterichA2026}) with
cross-study DDI validation (right).
GAP-1 and GAP-3 structural limits are confirmed by consistent prediction failure
across all three architectures.
MATCH cases are correctly detected.
AE: Adverse Effect. OAT: Organic Anion Transporter.
SSRI: Selective Serotonin Reuptake Inhibitor.}
\label{fig:taxonomy}
\end{figure}

Figure~\ref{fig:taxonomy} shows the Gap Taxonomy classification protocol from
the companion toxicity study \citep{DieterichA2026} alongside DDI prediction outcomes from this work.
The triangulation is structurally compelling: GAP-1 (serotonin pathway,
sertraline) and GAP-3 (renal OAT competition, probenecid) limits predicted by
the companion taxonomy are confirmed empirically by consistent failure across
all three DDI architectures.
MATCH cases (ibuprofen $P$\,=\,0.912, warfarin $P$\,=\,0.986) are correctly
detected, consistent with their structural encodability.

\section{Discussion}

\subsection{Why Cross-Attention Enables Mechanism Classification}

The asymmetric benefit of cross-attention reflects fundamentally different
information requirements.
Binary detection benefits from global molecular similarity: co-occurrence
patterns allow concatenated embeddings to encode implicit interaction likelihood.
Mechanism-type classification requires identifying which specific atoms mediate
the interaction --- information only accessible when atoms from molecule A attend
to atoms from molecule B before pooling.

The cross-attention patterns for ASA reference pairs (Section~3.5) support this
interpretation: pharmacophore-relevant atoms consistently receive highest
attention.
A secondary question is whether binary prediction quality depends on molecular
size: larger molecules provide more atoms contributing to the mean-pooled
embedding, potentially diluting pharmacophore signals.
This may partly explain why methotrexate (33 heavy atoms) is correctly
identified in binary mode despite its renal mechanism.

\subsection{Why the Ternary Architecture Fails}

The Ternary architecture's poor multi-class performance is consistent with a
training instability hypothesis.
The interaction graph edges are constructed from Top-3 cosine similarity between
post-cross-attention atom embeddings, creating a potential circular dependency.

Two promising directions emerge.
First, two-stage training: train the siamese MPNN and cross-attention to
convergence (as in CrossAtt), freeze those weights, then train only the
interaction graph MPNN.
Second, physically motivated inter-molecular edges --- based on hydrogen bond
donor/acceptor complementarity, electrostatic potential, or 3D distance criteria
--- would provide more chemically grounded training signal.

\subsection{Structural Limits and Pharmacovigilance Implications}

The consistent failure on sertraline and probenecid provides cross-validated
evidence for structural prediction limits: serotonin pathway and OAT transporter
mechanisms are outside the representational capacity of 2D structure-based GNNs,
independent of architectural complexity.

For pharmacovigilance practice under GVP Module IX, mechanistic plausibility is
required as a signal assessment triangulation element.
Structure-based GNN models can provide this plausibility for pharmacokinetic
and pharmacodynamic interactions mediated by structural features --- but not for
transporter-mediated or neurotransmitter pathway interactions.
The MNAR observation for paracetamol and vitamin C warrants attention: both have
clinically documented interactions with ASA absent from SSI-DDI, representing a
data annotation gap analogous to GAP-2 from the companion toxicity study \citep{DieterichA2026}.

\subsection{Limitations}

Synthetic negatives may underestimate true negative prevalence (MNAR).
No stratification by drug identity is applied in the train/test split; a
drug-level split would provide stricter generalization evidence beyond the
ASA cold-drug held-out set.
The curated reference set of seven pairs is too small for statistically robust
mechanistic conclusions.
The 86 SSI-DDI type codes lack standardized ontology mapping, limiting direct
clinical interpretation.
No 3D conformational information is used.
Cross-attention atom weights are reported for qualitative interpretation only.

\section{Conclusion}

This ablation study demonstrates that cross-attention between molecular graph
atoms is the critical architectural component for DDI mechanism-type
classification, while providing minimal benefit for binary interaction detection.
The $+45\%$ improvement in F1-macro from Concat to CrossAtt, combined with only
$+1.3\%$ improvement in binary AUC, establishes a clear design principle:
clinical applications requiring mechanism identification require cross-attention;
applications requiring only interaction screening may use simpler concatenation
architectures.

The central finding: \emph{cross-attention is necessary for mechanism-type
classification but largely unnecessary for binary DDI detection.}

The Ternary architecture's failure confirms that explicit interaction graphs
require stable pre-trained embeddings as initialization.
The consistent failure cases (sertraline, probenecid) provide cross-validated
evidence that transporter-mediated and neurotransmitter pathway interactions
represent a structural limit independent of architectural complexity ---
consistent with the companion toxicity study \citep{DieterichA2026}.

\section*{Conflicts of Interest}
The author declares no conflicts of interest.
No external funding was received for this study.

\section*{Data and Code Availability}
The SSI-DDI dataset is publicly available \citep{Nyamabo2021}.
Model scripts and result files are available from the author upon reasonable
request.
This work is licensed under CC BY 4.0
(\url{creativecommons.org/licenses/by/4.0}).

\clearpage
\section*{Glossary of Abbreviations}

\noindent{\small\textbf{Table 5:} Abbreviations used in this paper,
listed alphabetically.}\label{tab:glossary}

\vspace{6pt}
\noindent{\small
\begin{tabular}{ll}
\toprule
\textbf{Abbreviation} & \textbf{Full Term} \\
\midrule
ADR    & Adverse Drug Reaction \\
ASA    & Acetylsalicylic Acid (Aspirin; INN: ASA) \\
AUC    & Area Under the ROC Curve \\
BCE    & Binary Cross-Entropy (loss function) \\
COX    & Cyclooxygenase (COX-1 = PTGS1; COX-2 = PTGS2) \\
CYP    & Cytochrome P450 \\
DDI    & Drug-Drug Interaction \\
GNN    & Graph Neural Network \\
GVP    & Good Pharmacovigilance Practice (EMA guidelines) \\
MLP    & Multi-Layer Perceptron \\
MNAR   & Missing Not At Random \\
MPNN   & Message Passing Neural Network \\
NNConv & Neural Network Convolution \\
OAT    & Organic Anion Transporter \\
PD     & Pharmacodynamic \\
PK     & Pharmacokinetic \\
RDKit  & Open-Source Cheminformatics Library \\
ReLU   & Rectified Linear Unit (max(0,x)) \\
ROC    & Receiver Operating Characteristic \\
SMILES & Simplified Molecular Input Line Entry System \\
SSI-DDI & Substructure-Substructure Interaction DDI benchmark \citep{Nyamabo2021} \\
SSRI   & Selective Serotonin Reuptake Inhibitor \\
\bottomrule
\end{tabular}
}

\clearpage
\bibliographystyle{unsrt}
\bibliography{from_detection_to_mechanism}

@article{Pirmohamed2004,
  author  = {Pirmohamed, M. and others},
  title   = {Adverse drug reactions as cause of admission to hospital},
  journal = {BMJ},
  volume  = {329}, number = {7456}, pages = {15--19}, year = {2004}
}

@article{Bronstein2017,
  author  = {Bronstein, M.M. and Bruna, J. and LeCun, Y. and Szlam, A. and Vandergheynst, P.},
  title   = {Geometric deep learning: Going beyond {E}uclidean data},
  journal = {IEEE Signal Processing Magazine},
  volume  = {34}, number = {4}, pages = {18--42}, year = {2017}
}

@inproceedings{Gilmer2017,
  author    = {Gilmer, J. and Sch{\"u}tt, S.S. and Dahl, G.E. and Vinyals, O. and Riley, P.},
  title     = {Neural message passing for quantum chemistry},
  booktitle = {Proc.\ 34th ICML}, pages = {1263--1272}, year = {2017}
}

@article{DieterichA2026,
  author  = {Dietrich, J.},
  title   = {What Molecular Structure Cannot Tell Us: A Taxonomy of Explainability Gaps in {GNN}-Based Drug Toxicity Prediction},
  journal = {arXiv preprint arXiv:2605.26183},
  url     = {https://arxiv.org/abs/2605.26183},
  year    = {2026}
}

@article{Nyamabo2021,
  author  = {Nyamabo, A.K. and Yu, H. and Shi, J.Y.},
  title   = {{SSI-DDI}: substructure--substructure interactions for drug--drug interaction prediction},
  journal = {Briefings in Bioinformatics},
  volume  = {22}, number = {6}, pages = {bbab133}, year = {2021}
}

@article{Feng2022,
  author  = {Feng, Y.H. and others},
  title   = {{DPDDI}: a deep predictor for drug--drug interactions},
  journal = {BMC Bioinformatics},
  volume  = {23}, pages = {1--14}, year = {2022}
}

@article{Chen2023,
  author  = {Chen, Y. and others},
  title   = {{DSN-DDI}: an accurate and generalized framework for drug--drug interaction prediction by dual-view representation learning},
  journal = {Briefings in Bioinformatics},
  volume  = {24}, number = {1}, pages = {bbac597}, year = {2023}
}

@article{Lin2022,
  author  = {Lin, X. and others},
  title   = {Multimodal network for drug--drug interaction prediction using multi-source drug information},
  journal = {BMC Bioinformatics},
  volume  = {23}, pages = {1--15}, year = {2022}
}

@misc{Landrum2006,
  author = {Landrum, G.},
  title  = {{RDKit}: Open-source cheminformatics},
  year   = {2006},
  howpublished = {\url{https://www.rdkit.org}},
  note   = {Accessed: 24 May 2026}
}

@inproceedings{Kingma2014,
  author    = {Kingma, D.P. and Ba, J.},
  title     = {Adam: A method for stochastic optimization},
  booktitle = {ICLR}, year = {2015}
}

@article{Wishart2018,
  author  = {Wishart, D.S. and others},
  title   = {{DrugBank} 5.0: A major update to the {DrugBank} database for 2018},
  journal = {Nucleic Acids Research},
  volume  = {46}, number = {D1}, pages = {D1074--D1082}, year = {2018}
}

@article{Rubin1976,
  author  = {Rubin, D.B.},
  title   = {Inference and missing data},
  journal = {Biometrika},
  volume  = {63}, number = {3}, pages = {581--592}, year = {1976}
}

\end{document}